\documentclass[a4paper, 10 pt, conference]{ieeeconf}

\IEEEoverridecommandlockouts                            

\usepackage{cite}
\usepackage{amsmath,amssymb,amsfonts}
\usepackage{algorithmic}
\usepackage{graphicx}
\usepackage{textcomp}
\usepackage[utf8]{inputenc} %
\usepackage[T1]{fontenc}    %
\usepackage{hyperref}       %
\usepackage{url}            %
\usepackage{booktabs}       %
\usepackage{nicefrac}       %
\usepackage{microtype}      %
\usepackage{lipsum}
\usepackage[T1]{fontenc}
\usepackage{subfigure}
\usepackage{gensymb}
\usepackage{epsfig} %
\usepackage{xcolor}
\usepackage{booktabs}
\usepackage{array}
\usepackage{multirow}
\usepackage{ifpdf}
\usepackage{xkeyval}

\begin{document}

\title{Active Vibration Fluidization for Granular Jamming Grippers}

\author{
Cameron Coombe$^{1,2}$,
James Brett$^{1}$,
Raghav Mishra$^{1,3}$, \\
Gary W. Delaney$^{1}$, 
and David Howard$^{1}$ 
\IEEEmembership{Member, IEEE}
\thanks{$^1$CSIRO, Brisbane, QLD 4069 Australia (e-mail: david.howard@csiro.au)}
\thanks{$^2$School of Electrical Engineering and Robotics, Queensland University of Technology, QLD 4000 Australia}
\thanks{$^3$School of Mechanical and Mining Engineering, University of Queensland, QLD 4070 Australia}
\thanks{Corresponding author: David Howard (e-mail: david.howard@csiro.au).}
}

\maketitle

\begin{abstract}
Granular jamming has recently become popular in soft robotics with widespread applications including industrial gripping, surgical robotics and haptics. Previous work has investigated the use of various techniques that exploit the nature of granular physics to improve jamming performance, however this is generally underrepresented in the literature compared to its potential impact.  We present the first research that exploits vibration-based fluidisation actively (e.g., during a grip) to elicit bespoke performance from granular jamming grippers.  We augment a conventional universal gripper with a computer-controllled audio exciter, which is attached to the gripper via a 3D printed mount, and build an automated test rig to allow large-scale data collection to explore the effects of active vibration. We show that vibration in soft jamming grippers can improve holding strength.  In a series of studies, we show that frequency and amplitude of the waveforms are key determinants to performance, and that jamming performance is also dependent on temporal properties of the induced waveform.  We hope to encourage further study focused on active vibrational control of jamming in soft robotics to improve performance and increase diversity of potential applications.

\end{abstract}

\begin{keywords}
Soft manipulation, Soft robotics applications, Vibration control 
\end{keywords}

\section{Introduction}
Soft robots use soft and compliant structures, exploiting the mechanical properties and non-linear behaviours of their constituent materials to provide adaptive behaviours.  Nowhere is this more prevalent than in the tasks of gripping and manipulation \cite{shintake2018soft}, where compliance and variable stiffness are harnessed to conform to irregular object shapes, maintaining larger contact surfaces and exerting fewer extraneous contact forces. 

Pneumatics are a standard approach to gripping, and provide this compliance through deformation. However, they struggle to exert strong grips \cite{xavier2022soft}. Many gripping tasks require both deformation and variable stiffness, to provide adaptability/compliance and grip strength respectively. As such, variable-stiffness soft structures are a common solution which adapt the component's compliance \cite{manti2016stiffening}. A popular implementation of a variable stiffness soft structure exploits the jamming phase transition, through e.g., laminar, fibre, or granular jamming \cite{fitzgerald_review_2020}. Granular jamming is the most common of the three, whereby granular materials are driven from fluid-like to solid-like states through the variation of (mainly frictional) inter-particle forces under vacuum pressure. Granular jamming designs are simple to create and actuate, and exhibit a wide range in possible stiffness values, rapid stiffness variation, and a wide array of possible shapes and sizes.

\begin{figure}[t!]
\centering
\includegraphics[width=0.297\columnwidth]{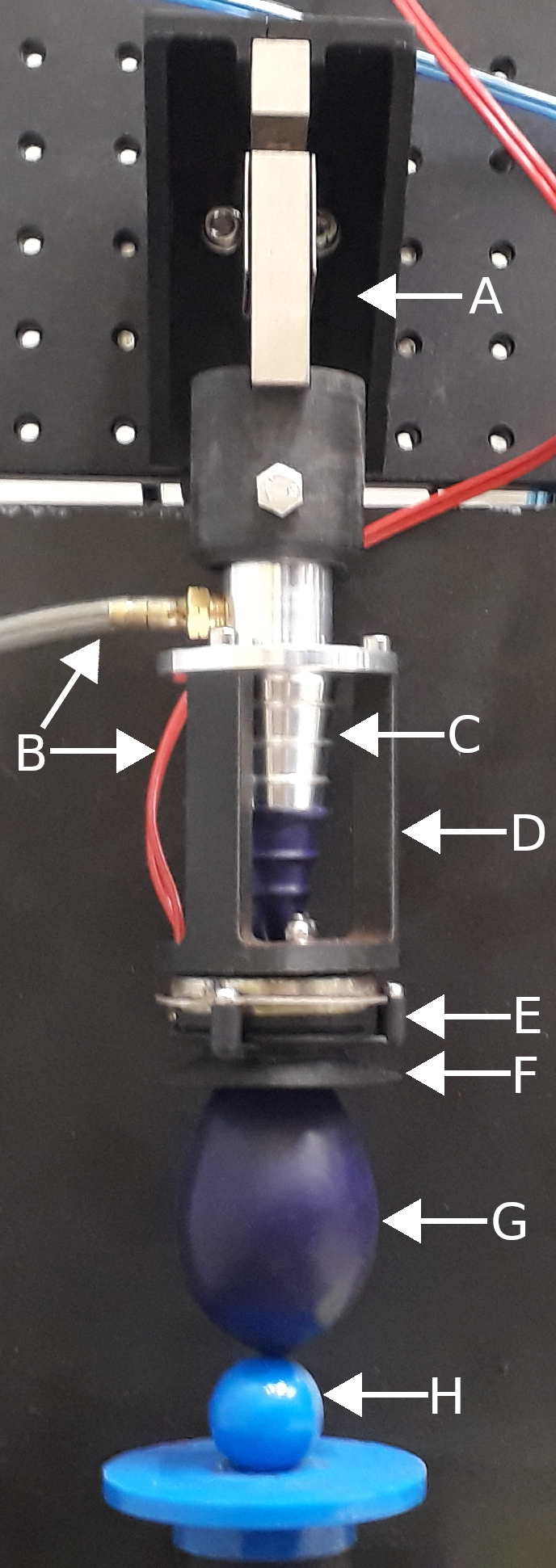}
\includegraphics[width=0.45\columnwidth]{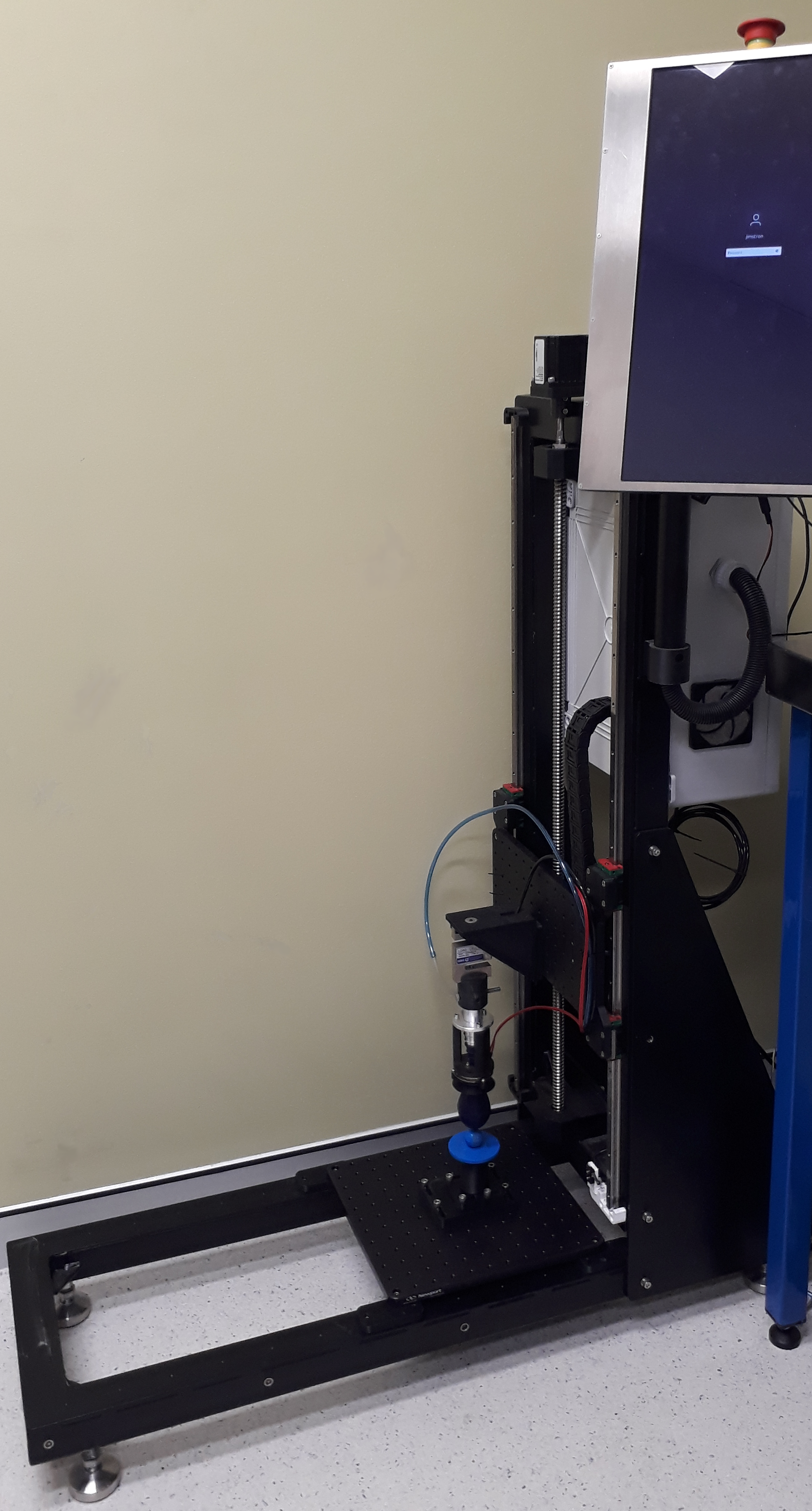}
\caption{Automatic vibration-based fluidising jamming gripper setup. (A) Load Cell (B) silicone tube (to vacuum pump) and electrical wiring (to amplifier) (C) gas outlet (D) 3D printed audio exciter mount (E) audio exciter (F) gripper shroud (G) latex membrane and coffee ground gripper (H) 3D printed target object and base.}
\label{fig:jimstron_photo}
\end{figure} 

The archetypal example of granular jamming for robotics is the granular jamming-based universal gripper \cite{brown_universal_2010}. Grains (e.g. ground coffee) are surrounded by a soft membrane and jammed by application of negative pressure from a vacuum pump after being pushed down and moulded around the target object. The resultant stress from the atmospheric pressure causes a jamming phase transition to occur; the gripper hardens and is able to maintain a grip via the interaction between the target object, the granular material, and the soft membrane. This design allows for robust 'universal' gripping that is suitable for a variety of objects and does not require software to solve complex manipulation planning problems, simultaneously offering a simple and generally-applicable gripping solution. 

Due to this potential, a number of previous studies have investigated methods to improve performance in jamming systems, typically by exploiting the underlying granular physics that governs the bulk behaviour of the gripper, e.g. membrane coupling \cite{jiang2012design} and actively varying the inner volume\cite{fujita2016variable}.

The most well-known of these phenomena is fluidisation, where the gripper is forced into a fluid-like state.  Traditional granular jamming grippers experience partial jamming while moulding over the target object.  As the gripper pushes down on the object and begins to conform, stress builds up within the gripper, causing jamming and reducing compliance.  This eventually prevents the gripper moving over the object. Vibration increases both compliance and the contact surface area during the moulding process, leading to greater holding forces. This process can change the push force exerted on the object during moulding by relaxing the stress built up in the grains allowing for them to either be compressed tighter, or maintain a relaxed grip.  In the literature is realised through the use of bursts of positive pressure \cite{amend_positive_2012}, which is shown to improve the speed of resetting the bulk granular state between grips, as well as improving grip strength.

Positive pressure is not the only way to fluidize, and in fact has numerous disadvantages.  First, it requires an extra vacuum pump and associated silicone tubing, which significantly increases the size and weight of the experimental apparatus.  Additionally, it does not support the ability to precisely control the fluidization effects - positive pressure is typically send in a 'bang-bang' fashion (positive pressure is either on or off, and nothing in between).

Here we introduce active vibrational jamming, showing vibration to be a simpler and more controllable solution that more readily supports the generation of diverse embodied behaviours.  Vibration of granular materials has been reported for  unjamming, in industries that require flow of granular materials such as food processing, pharmaceuticals, and agriculture \cite{delaney2012}.  It has been mentioned in the context of granular jamming, but only as a mechanism to reset the bulk state of the gripper between grips, rather than as an active component that augments the gripping process itself.  

We develop a novel setup where a small, programmable vibration element is mounted and placed in contact with the gripper membrane. We show that vibration improves holding force and influences the pre-loading forces exerted on the target object. We additionally show that different input vibration waveforms can influence holding forces, which provides further opportunity to optimise the gripper for particular objects and applications.  Experimentation uses a fully automated testing rig, allowing us to collect extensive data sets. 

\section{Background}

\subsection{Granular Jamming in Soft Robots}
The gripper displayed in this paper is a modification of the original universal jamming gripper \cite{brown_universal_2010}. Previous work has been performed on modifying the original design to push the grippers capabilities, such as adding positive pressure to counter the inability to grip traditionally difficult objects \cite{amend_positive_2012} and exploring force-feedback for closed loop control \cite{nishida_development_2014}. The exploration of these grippers capabilities have also been applied to different environments including underwater usage \cite{licht_universal_2016}, and a scaled-up variant used on a crane \cite{miettinen_granular_2019}.

Different material properties have been explored as a means to tune jamming performance, with both shape and size of grains playing a major role \cite{howard2021shape}, as well as multi-material membranes to induce programmed deformations \cite{howard_one-shot_2021}.  Evolutionary algorithms \cite{fitzgerald_evolving_2021} were used to generate bespoke grains for soft jamming grippers, as well as to explore the possible configuration space of granular morphologies in order to optimise properties for bespoke granular design \cite{delaney_multi-objective_2020}.  The exploitation of granular jammings variable stiffness properties have been explored in a plethora of morphologies beyond 'bag-style' universal gripper designs, including multi-finger grippers, paws for legged robots \cite{chopra_granular_2020,howard_one-shot_2021}, prostheses \cite{cheng_prosthetic_2016} and haptic feedback devices \cite{zubrycki_novel_2017}. Aside from positive pressure, there has been little work in exploring the utilisation of other granular physics effects in jamming-based robots.

Fluidisation has been covered in a few studies, e.g., through periodically \cite{amend_positive_2012} or continually \cite{kapadia_design_2012} modulating the air pressure  inside the membrane.  These setups require two vacuum pumps or elaborate pneumatic circuits to function, and offer only a single control variable to create high performance behaviours.  

Vibration is briefly mentioned \cite{amend_soft_2016}, where a vibration motor replaces a positive pressure pump to unjam/reset the gripper, rather than as an active element to enhance the grip.  Smaller vibration motors embedded in the granular material did not have enough power to unjam the gripper, while larger motors could not be embedded in the gripper without changing bulk properties.

Compared to the above work, our method of fluidisation by vibration is superior in many regards.  Mainly, our fluidisation is decoupled from application of vacuum, and is programmable with a wide range of input waveforms. We use a compact audio exciter as a vibration element mounted outside the gripper membrane. The exciter has a large bandwidth allowing for a large range of inputs, unlike vibration motors which need to be tuned to specific frequencies.  Importantly, ours is the first work to exploit vibration during a grip.  Our method is mechanically simpler as it does not rapidly alternate pump flow through a valve, which requires 2-way valves and a controller or two pumps, and causes extra fatigue cycles in the gripper and supporting equipment.  

\subsection{Vibration in Jamming Systems}
\label{vibra}

Although vibration-based fluidisation has not been used actively in soft robotics, it is well-studied within the granular physics literature.  Vibratory Fluidised Beds are commonly used for unjamming in industries dealing with food, powder, pharmaceutical and agricultural processing, e.g., to enable smooth flow down a hopper \cite{janda_unjamming_2009}. 

Much work has been done trying to understand the origins of granular material properties on the mesoscopic scale, using 2D lattice gas automata models for studying vibration in granular materials \cite{nicodemi_compaction_1997, nicodemi_frustration_1997}, showing properties such as glass transition and density relaxation.   Granular materials agitated using vibrational beds show different internal energies depending on the input frequencies, even if the energy corresponding to the vibrational system is kept constant, due to the resonant modes of the bulk material\cite{windows-yule_resonance_2015}.  Discrete Element Method (DEM) simulations of granular matter under vibration showed the presence of linear vibrational modes at low amplitudes with bulk elastic modulus softening behaviour, which was attributed to the remaking of internal force chain contacts during vibration \cite{lemrich_dynamic_2017}.

\subsection{Summary}
Overall, placing our work in the context of the wider literature shows several novel contributions.  We present the first jamming gripper that leverages vibration as a means of active fluidisation and utilise this during the grip operation, achieving a programmable output.  Vibration is shown to increase grip strength, influence stress during object moulding, and increase surface contacts between gripper and object.  We use our flexible audio exciter-based setup to run experiments with a variety of input waveforms.

\section{Methodology}
 \subsection{Materials and Design}
The vibration-enhanced gripper is similar to a traditional granular jamming gripper \cite{brown_universal_2010}, with the addition of an audio exciter which agitates the soft membrane and grains from above (Fig:  \ref{fig:fig2a}).  The gripper contacts the exciter as it is pushed down transferring more energy as it moulds around the target object. A 3D printed shroud between the balloon and exciter facilitates the moulding around the object as well as distributing the forces exerted on the exciter to protect it. All 3D printed components were printed using a Stratasys Fortus 370 printer unless otherwise specified.

Uninflated 27.5cm (11 inch) Qualatex latex balloons were used as soft membranes. These were filled with 27g of coffee grounds to the base of the neck. Party balloons are easily sourced and low cost however are far from ideal gripper membranes.  Balloons lasted between 25 and 120 tests. This is primarily caused by the stress imparted on the neck during pull off which slowly weakens the material until small holes appear. Balloons were replaced after breakage. 

For our application, we must vibrate the gripper with enough power to cause fluidisation of the granular material, without hampering either its bulk properties or its mechanical compliance. Additionally, it is desirable if the vibration element allows control over input waveforms and oscillation frequency. Audio speaker exciters have a small footprint and large bandwidth to accommodate most of the human hearing range from 20Hz to 20kHz within its pass-band making them a desirable choice. We use the Tectonic 30W TEAX32C30 Speaker-based exciter drivers, which were found to output enough power to cause fluidisation, while having a suitable bandwidth to allow for controllability. A hole is drilled through the exciter magnet for the balloon neck to pass through and attach to the gas outlet. We measured the full range of used output frequencies to ensure this had not altered the frequency response.
 
 \subsection{Experimental Setup}

\begin{figure}[h!]
\centering
\includegraphics[width=0.95\columnwidth]{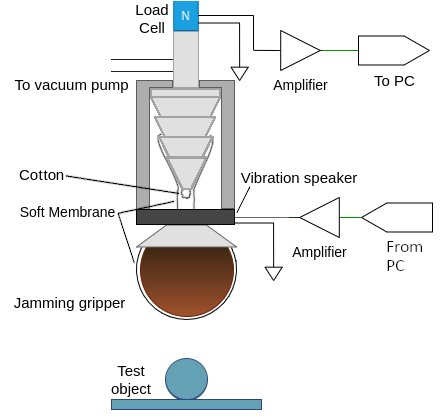}
\caption{Simplified diagram of the test setup used for the experiments in the paper, showing the gripper connected to the UTM and the test object.}
\label{fig:fig2a}
\end{figure}

Our automated Universal Testing Machine (UTM) (Fig: \ref{fig:jimstron_photo}) consists of a controllable z axis. Grippers are attached via a 3D printed mount. The gripper apparatus (Fig: \ref{fig:fig2a}) consists of a Zemic H3-C3-25kg-3B load cell, connected to an Intel NUC through a a Wondom AB13223V130 amplifier, inline with the audio exciter and gripper. The NUC controls the z axis, balloon pressure valve and audio exciter. A 3D printed adapter attaches the exciter to the gas outlet. The outlet is filled with cotton to prevent coffee flowing out of the membrane, and to avoid damage to the neck of the balloon during jamming. The cotton is periodically replaced.  Vacuum is provided by a Rocker 300 vacuum pump.

A spherical target object 25mm in diameter was 3D printed on a Stratsys Objet 500 Connex3 printer to test the holding strength of our gripper (Fig.\ref{fig:fig2a}). This size was chosen as the gripper diameter was approximately 45mm and traditional jamming grippers are known to display degraded performance for objects larger than about half the size of the gripper \cite{brown_universal_2010}.   A larger 70mm diameter circular base was also printed to simulate picking the object up off a floor.

\subsection{Experiment Procedure}
We performed three sets of experiments varying push down height, frequency and volume (Tab: \ref{tab:experiment_summary}. The latter two experiments are separated into time invariant and time varying waveforms. For each unique condition we performed 20 tests with a separate 84 height tests resulting in 746 total tests. Additionally due to manufacturing differences balloons have different baseline grip strengths. Therefore there is a trade off between number of tests per condition per balloon and number of balloons per experiment. We have opted to use at least 3 balloons per experiment/waveform temporal type with 10 tests per condition per balloon. More balloons are used if needed. This means that for the single tone volume experiment, ideally 3 balloons were tested 70 times, or 10 times per volume level. Within the course of a single balloon, all possible test conditions are cycled through in a random order then cycled through again until 10 cycles had been complete or the balloon breaks. This avoided any bias from a single test condition always occurring near the end of a balloons lifecycle.

Each grip test begins by zeroing the load cell and applying 0.3 and 0.01 seconds of pre-grip vacuum and positive pressure respectively to remove excess air from the balloon. The balloon starts positioned 100mm above the test object base and is lowered at 30mm/s to the push down height of 30mm above the test object base while the audio file plays through the audio exciter. Once the push down height is reached the audio stops and the vacuum is applied. After 10 seconds of vacuum pressure the gripper is lifted at 30mm/s until 70mm above the test object base. Three pulses of positive pressure for 0.3 seconds are applied and the secondary 200Hz vibration signal plays to release the object if it is still gripped due to interlock forces. This avoids damage caused by stretching the balloon neck. The gripper is then raised to 100mm above the test object base and three more 0.3 second positive pressure pulses are applied to unjam it. 

\begin{table*}[t!]
\caption{Summary of all experiments conducted. Sweeps refer to 1, 25 second chirp while pulses refer to 25, 1 second chirps played in succession. 20 randomly sampled tests for each variable increment were selected from each population to be used in analysis. }
\label{tab:experiment_summary}
\begin{tabular}{|l|l|lll|lll|}
\toprule
Experiment & Height & \multicolumn{3}{l|}{Frequency} & \multicolumn{3}{l|}{Volume} \\ \cline{2-8} 
Temporal Type & Invarient & \multicolumn{1}{l|}{Invarient} & \multicolumn{2}{l|}{Varying} & \multicolumn{1}{l|}{Invarient} & \multicolumn{2}{l|}{Varying} \\
Waveform Type & Tone & \multicolumn{1}{l|}{Tone} & Sweep & Pulse & \multicolumn{1}{l|}{Tone} & Sweep & Pulse \\
Height (mm) & 27-69 & \multicolumn{1}{l|}{29} & \multicolumn{2}{l|}{29} & \multicolumn{1}{l|}{29} & \multicolumn{2}{l|}{29} \\
Frequency (Hz) & 200 & \multicolumn{1}{l|}{\begin{tabular}[c]{@{}l@{}}10, 25, 50, 100, 150, 200, \\ 300, 400, 500, 600, 700, 800\end{tabular}} & \multicolumn{2}{l|}{\begin{tabular}[c]{@{}l@{}}1-100, 100-1, 100-200, 200-100, \\ 100-400, 400-100, 100-800, 800-100\end{tabular}} & \multicolumn{1}{l|}{200} & \multicolumn{2}{l|}{200} \\
Volume (\%) & 0, 150 & \multicolumn{1}{l|}{150} & \multicolumn{2}{l|}{150} & \multicolumn{1}{l|}{\begin{tabular}[c]{@{}l@{}}0, 25, 50, 75, \\ 100, 125, 150\end{tabular}} & \multicolumn{2}{l|}{\begin{tabular}[c]{@{}l@{}}75-150, 150-75, 0-150, \\ 150-0, 0-75, 75-0\end{tabular}} \\
n Balloons & 1 & \multicolumn{1}{l|}{5} & 3 & 3 & \multicolumn{1}{l|}{3} & 3 & 3 \\
n Valid Tests & 84 & \multicolumn{1}{l|}{331} & 180 & 234 & \multicolumn{1}{l|}{210} & 162 & 145 \\
n Sampled Tests & 84 & \multicolumn{1}{l|}{240} & 160 & 160 & \multicolumn{1}{l|}{140} & 120 & 120 \\
\bottomrule
\end{tabular}
\end{table*}

\subsection{Analysis}
\begin{figure}
\centering
\includegraphics[width=0.3\textwidth]{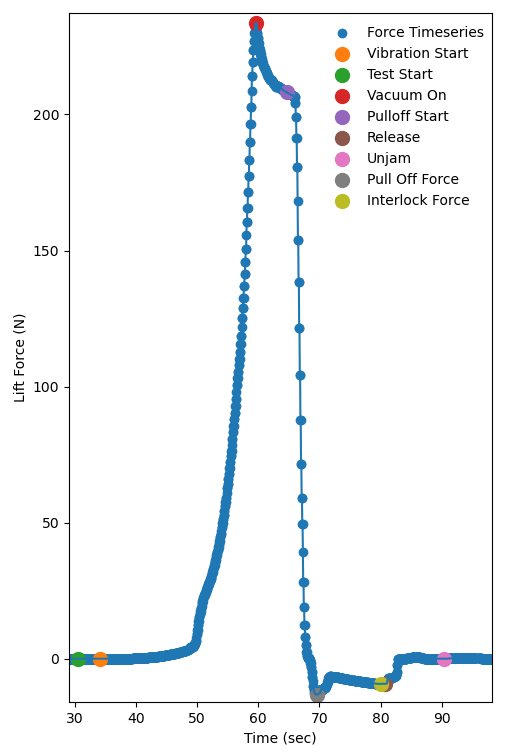}
\caption{\label{fig:force_timeseries}Typical force progression of a grip test where the gripper interlocks with the object. Push force is taken as the maximum of this graph. Holding force is defined as the first valley after the force crosses 0 with a preceding gradient $>$ 1 and interlock is the last valley before release with a preceding gradient $<$ 1.}
\end{figure}

Two force measurements are presented: holding and push force (Fig: \ref{fig:force_timeseries}). Push force is defined as the highest peak of force seen during a single grip test, always occurring just before the vacuum is applied. The holding force was taken as the first peak after crossing 0 due to pull off excluding the point marked "Interlock Force" in Fig: \ref{fig:force_timeseries}. Mann-Whitney U tests are used to compare each pair of conditions within each experiment/waveform temporal type, with significance assessed at P$<$0.05.

\section{Results}

\subsection{Time Invariant Waveforms}

We first quantify the effect of time invariant vibration signals on gripper behaviour.  We conduct an experiment comparing the holding strength across different input vibration frequencies and volumes. The frequency is varied first from 10Hz to 200Hz in small intervals then from 200Hz to 800Hz in intervals of 100Hz. Volume is increased in intervals of 25\% from 0-150\%. 

\begin{figure}[h!]
\centering
\includegraphics[width=0.95\columnwidth]{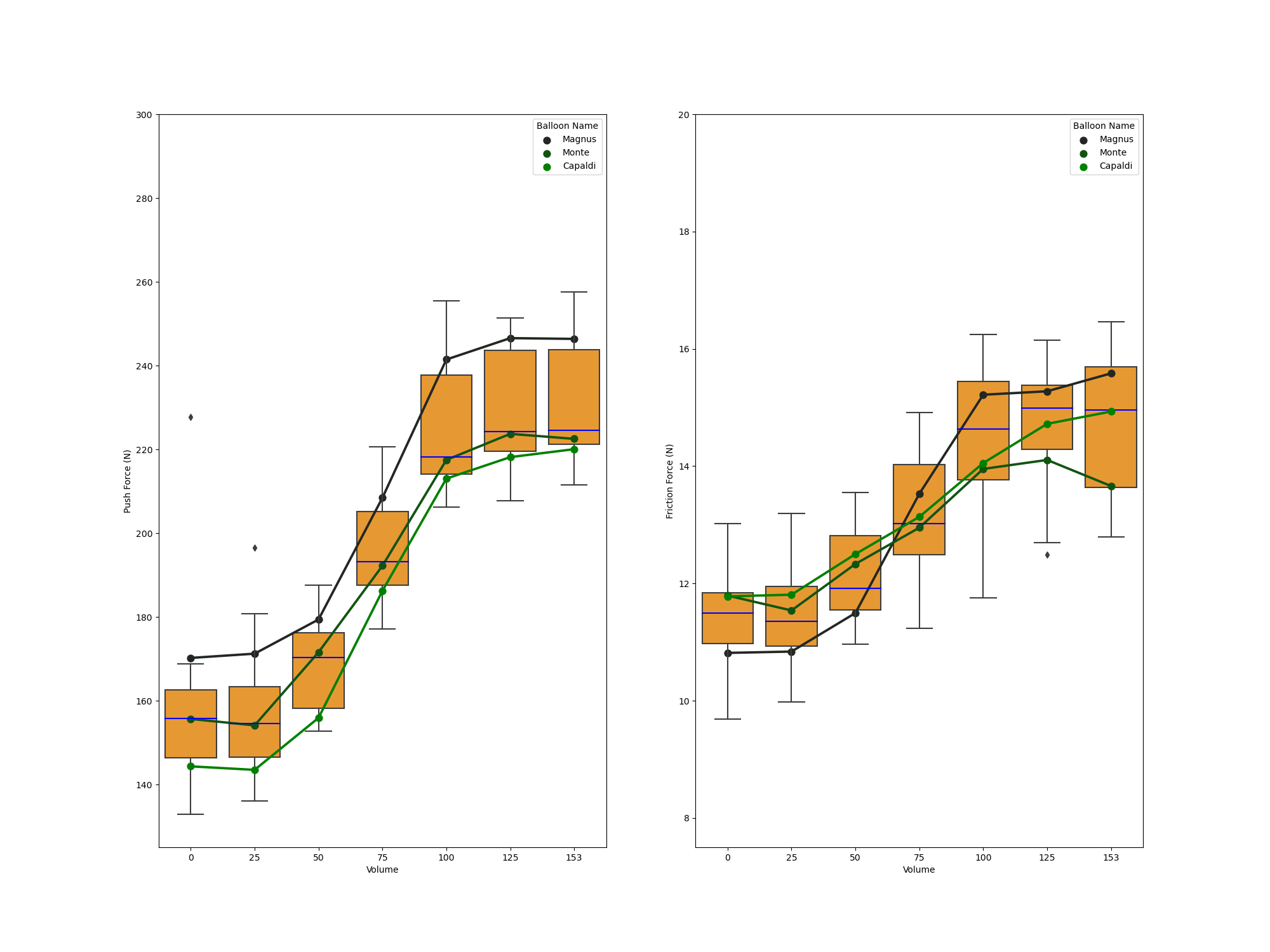}
\caption{Push down (left) and holding (right) forces for time invariant volume tests showing an increase in grip strength with an increase in volume.}
\label{fig:VolTone}
\end{figure} 

\begin{figure}[h!]
\centering
\includegraphics[width=0.95\columnwidth]{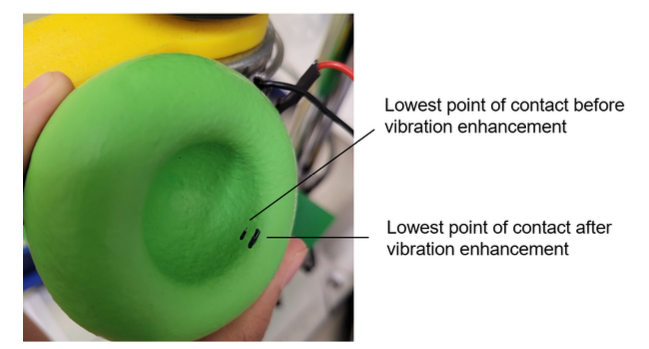}
\caption{Vibration applied to a jamming gripper on 30mm sphere object achieves a lower point of contact when gripped, leading to a larger contact area for gripping}
\label{fig:relaxation_photo}
\end{figure} 

\begin{figure}[h!]
\centering
\includegraphics[width=0.95\columnwidth]{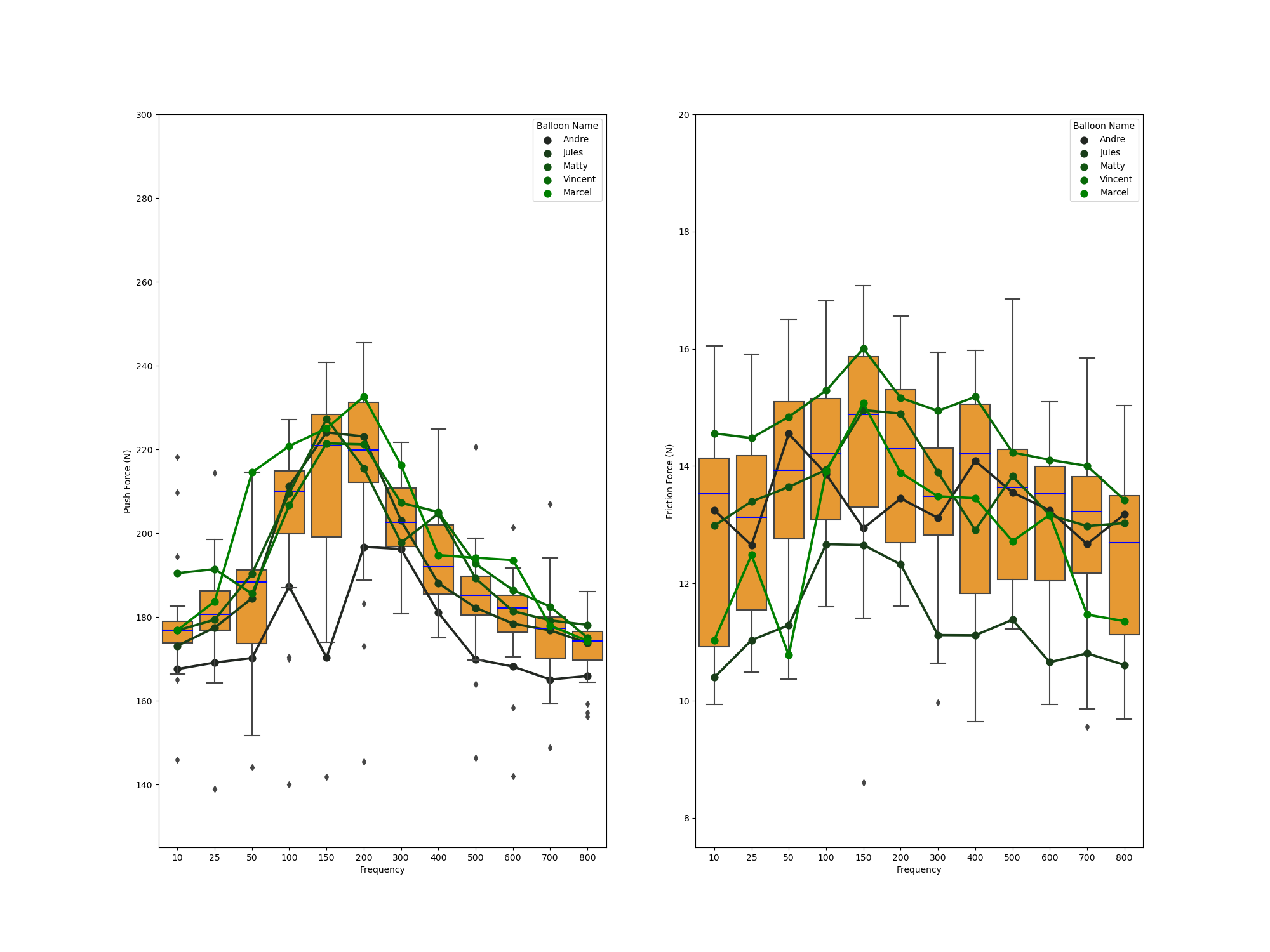}
\caption{Push down (left) and holding (right) forces showing that frequencies surrounding 150Hz provide the strongest holding forces.}
\label{fig:FreqTone}
\end{figure} 

Increasing volume shows a trend of increasing holding and push force from an average of 12N to 14N (p < 0.05) at maximum and from 150N to 250N for push force (Fig: \ref{fig:VolTone}). Results from all statistical tests can be found in the appendix. The trend approximates a sigmoid curve, indicating a cap on the improvements in grip strength that can be gained through excitement. We can explain these results in terms of improved gripper-object contact.  We perform a single test and mark the edge of the object impression with a marker after the gripper comes to a stop before vibration and after the gripper has experienced vibration. As shown in Figure \ref{fig:relaxation_photo}, the marks are $\approx$2.5mm apart, indicating an improved contact angle and increased enveloping of the object. This improved contact angle results in an approximately O-ring strip of additional contact area, providing a larger area for increased grain-object contacts and stronger static friction effects.

When varying frequency, a clear trend in holding force is observed, increasing from a mean of 13N  at 25Hz to 15N at 150Hz (p < 0.05) then decreasing to 12.5N at 800Hz (p < 0.05) (Fig: \ref{fig:FreqTone}). Push force follows a similar trend from 175N to a maximum of 225N. We propose that this is due to the jamming gripper having low frequency fundamental modes which are not as activated by higher frequencies. Additionally, it is well known that lower frequency modes within granular materials have higher participation ratios i.e. the modes create more global involvement of the grains in vibration, while higher frequency modes dissipate energy mostly into the part of the material that is close to the vibration element \cite{chen_low-frequency_2010}. Results indicate that a lower frequency is beneficial, as the higher participation would imply that force chains in the lower half of the gripper are broken away from the vibration element more easily.  This also opens up the possibility to target higher frequency localised vibrations into specific regions of a jamming structure, however such experimentation is out of scope for our current study.

\subsection{Time Varying Waveforms}

Our second experiment aimed to observe if there are any effects on the pull force as a result of time-dependent frequency and amplitude variations in the input waveform. These time dependent effects would change the behaviour as high amounts of vibration allow more random movement and increased rearrangements of force chains.  Conversely low frequency vibrations allow grains more time to relax by providing them sufficient energy and time to explore lower energy configurations. Finally, a lack of vibrations leads to freezing of the current state of the material due to the inherently high damping ratios in granular materials.  We can analyse the vibrational response of our granular material in terms of the modes that are activated. Like all structures, granular structures have natural vibrational modes. Vibratory inputs of different frequencies will activate different modes which can vary in their properties and cause local energetic interaction variations over the volume of the gripper such as the previously-described localisation of high frequency modes.

\begin{figure}[h!]
\centering
\includegraphics[width=0.95\columnwidth]{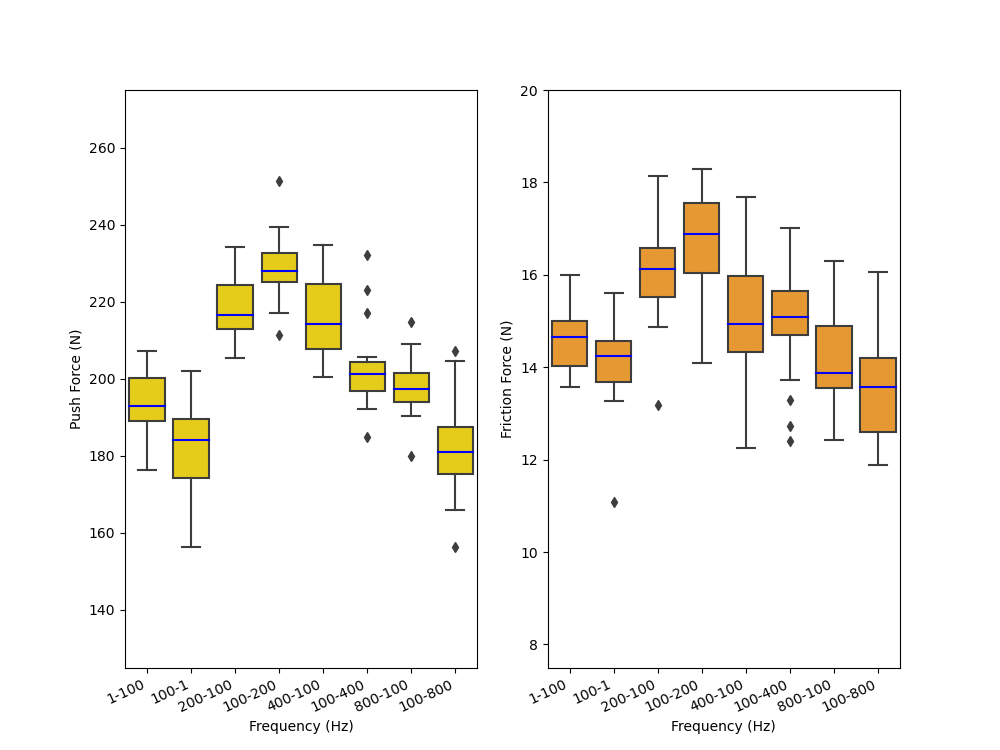}
\includegraphics[width=0.95\columnwidth]{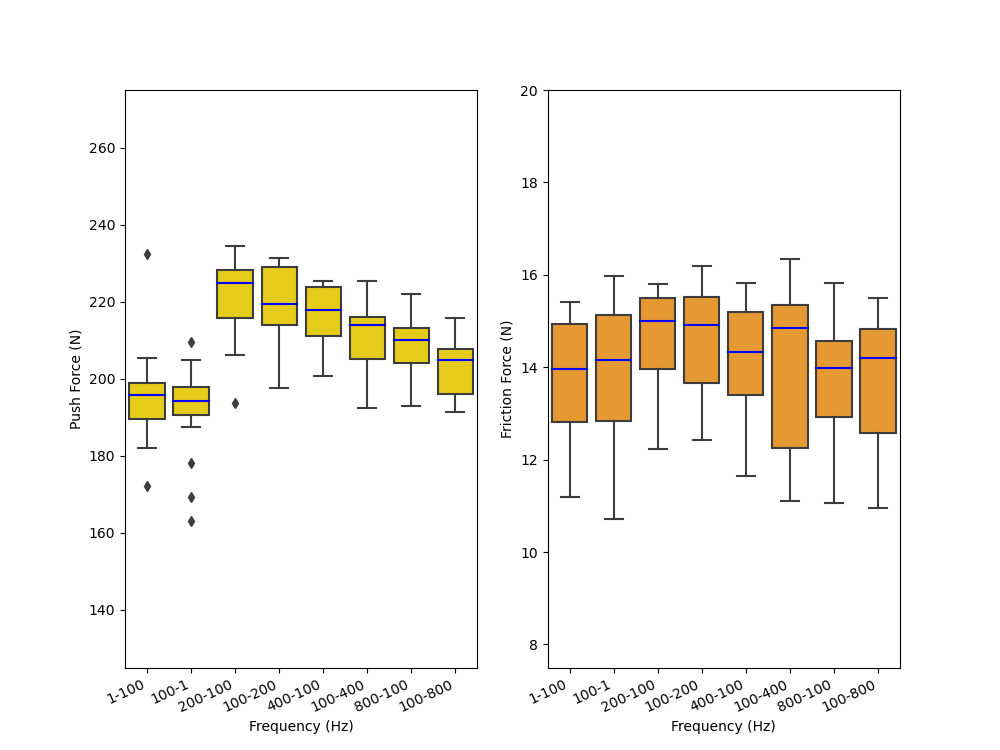}
\caption{Push down (left) and holding (right) forces for time varying frequency tests. (Top) 25 second chirp signals or 'sweeps' across a range of start and stop frequencies show higher holding forces for frequencies around 150Hz. (Bottom): 25 1 second chirp signals or 'pulses' show little variation.}
\label{fig:FreqSweepPulse}
\end{figure}

A myriad of experiments were conducted with varying vibration input sine waveforms in the form of ramps up and down in frequency (chirps) and amplitude. Specifically, the input frequency was ramped (see Fig.\ref{fig:FreqSweepPulse}) to and from 100Hz for 1Hz, 200Hz, 400Hz and 800Hz. Lastly, the frequency was held constant at 200Hz while the amplitude was ramped to and from each combination of 0, 75 and 150\% intervals. Two variations of these experiments were run varying ramp duration, the first with 1, 25 second ramp or 'sweep' and the other with 25, 1 second ramps or 'pulses' played in succession while the gripper descended.

\begin{figure}[h!]
\centering
\includegraphics[width=0.95\columnwidth]{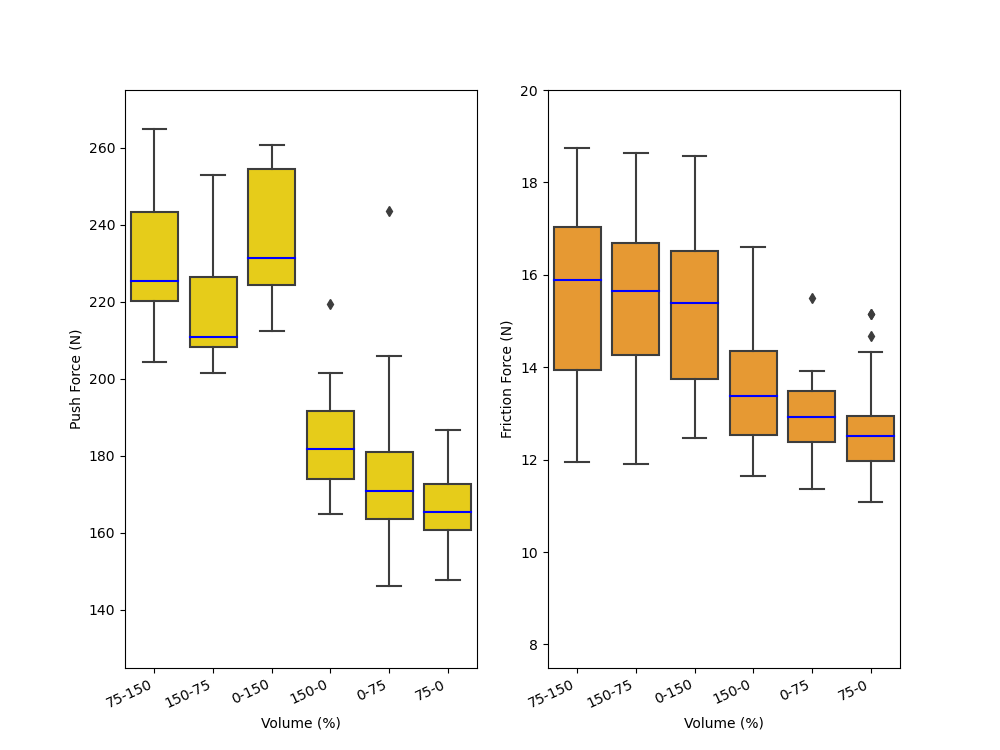}
\includegraphics[width=0.95\columnwidth]{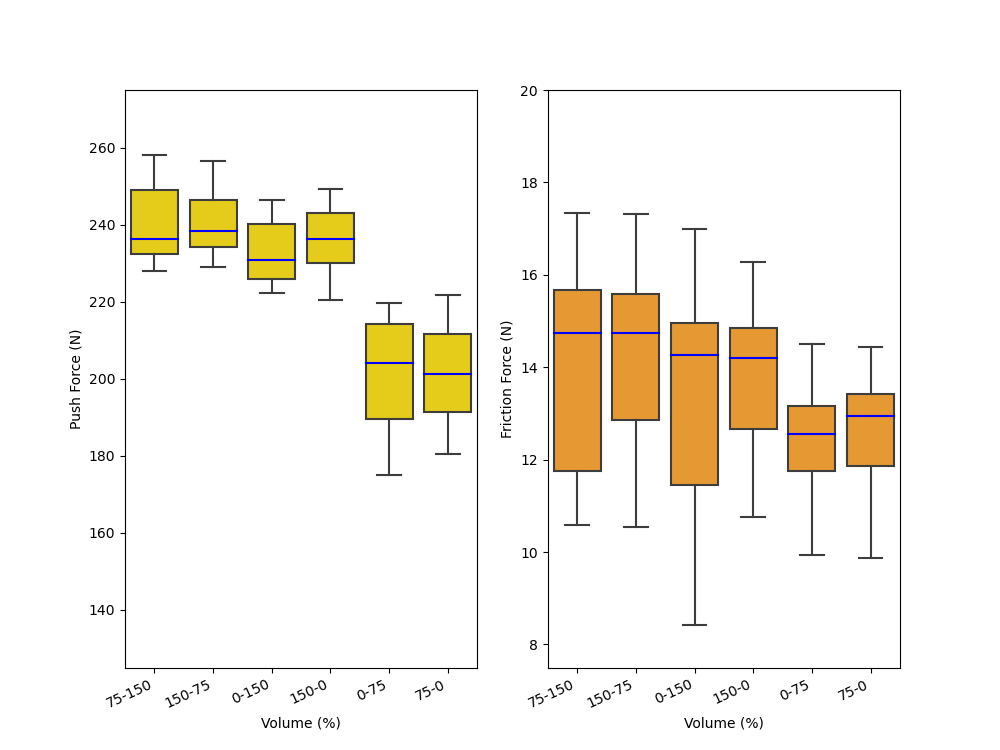}
\caption{Push down (left) and holding (right) forces for time varying volume tests. (Top) 25 second chirp signals across a range of start and stop volumes show higher holding forces for higher volumes overall and during the latter half of the test. (Bottom): 25 1 second chirp signals or 'pulses' show higher forces for higher volumes.}
\label{fig:VolSweepPulse}
\end{figure} 

Results demonstrate a clear dependency between temporal properties of the input signal and gripper behaviour. Often time varying waveforms outperform even the best constant tones. 100-200Hz (17N), 75-150\% (16N), 150-75\% (16N), 0-150\% (16N) frequency and volume sweeps are seen to have very high grip strengths, and specifically display higher holding force than a constant 200Hz vibration. In some cases the result is easily explained, as slowly ramping down from high to low vibration allows the granular material to settle into a more compact and stable configuration in a process similar to annealing. 
The reason for ramping volume producing high pull forces is possibly due to a window for grain relaxation occurring later in the grip after it has partially moulded over the balloon. This is bidirectional in the case of 75 and 150\% but not for 0 and 153\% tests.  
The pulses show a trend seen in the time invariant waveform data where waveforms that spend more time around better performing frequencies or higher volumes perform better.%

\subsection{Gripper Relaxation}

\begin{figure}[h!]
\centering
\includegraphics[width=0.95\columnwidth]{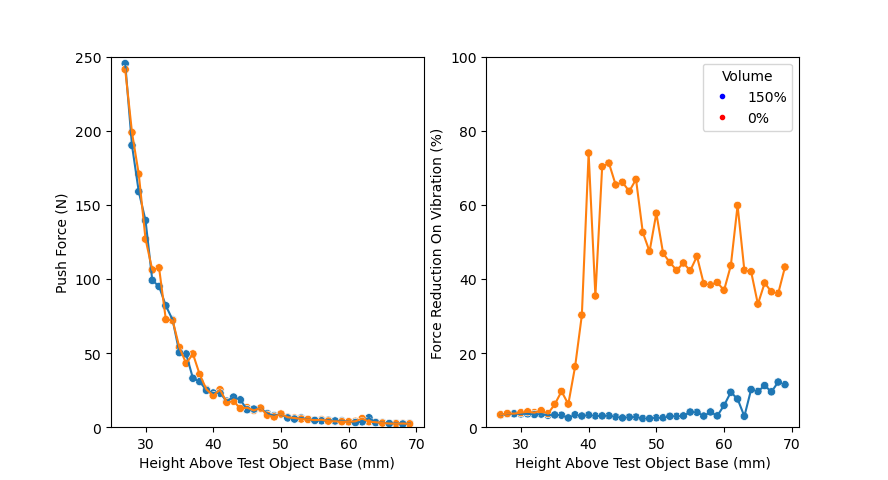}
\caption{Absolute push force (left) and the relaxation caused by vibration (right) showing that vibration causes a reduction in push force during the initial stages of gripping.}
\label{fig:RelaxationGraph}
\end{figure}

To assess the relaxation effect of vibration, two tests were run for each 1mm push down height increment between 27-70mm above the test object base. This ranges from the maximum to slightly above the minimum grip height. For one of these two tests, the exciter was fed a 5 second 200Hz sine wave after the push down height was reached. For a small range of heights these vibrations result in rapid stress relaxation within the gripper as the particles rearrange to pack better around the target object (Fig.\ref{fig:RelaxationGraph}), leading to a rearrangement of the force network within the granular material and an overall reduction in downwards force on the target object of $\approx$71\%.

It is generally useful to minimise the force required to grip an object, but is especially desirable when the object is soft or fragile as it would cause less deformation or fewer forces to contribute to fracture respectively. Note that for Fig.\ref{fig:RelaxationGraph}, the vacuum pump was not turned on for the gripping action as it is not necessary to show the stress relaxation.  In other words, we are measuring push force only, not pull force.

\section{Conclusion}

Despite the extensive exploration of granular jamming universal grippers, robotics literature has largely failed to exploit the key results on granular material behaviour that exist in the field of condensed matter physics. In this work, we have exploited one result applying vibration to fluidise granular material. We attached an audio speaker exciter to the membrane of the jamming gripper from above which transmits vibrations through the membrane to the granular material. The resulting gripper maintained higher holding forces on the spherical test object, presented larger contact surface area between object and gripper however also resulted in higher push down forces in all but certain cases. We tested the gripper with a variety of input frequencies, showing that grip strength favours lower frequencies and higher volumes. More complex chirp and amplitude ramp waveforms strongly increase holding force more than single tones indicating more complex behaviour. Further work is needed to understand how the time-varying vibration parameters affect the granular material inside the gripper and its resultant holding force, especially for complex waveforms.  

There are significant opportunities for future research in bringing in other granular matter physics effects and theory and using them to improve jamming grippers or create other novel jamming structures. Possible avenues may include alternative ways of jamming; studying granular flow and wave propagation effects; exploring the relationship between the size of the object, frequency of oscillation and the size of the gripper; and exploiting crystallisation and compaction effect, in particular for granular materials with specified particle morphologies \cite{delaney2010g}. Many effects will require experimentation to determine, and have the potential to generate significant theoretical as well as practical contributions. The field offers a rich vein of opportunities for both fundamental and applied research. We hope to encourage others to consider these possibilities and explore how they may improve or introduce new granular matter-based soft structures for robotics. 

\bibliographystyle{IEEEtran}
\bibliography{references}

\begin{thebibliography}{10}
\providecommand{\url}[1]{#1}
\csname url@samestyle\endcsname
\providecommand{\newblock}{\relax}
\providecommand{\bibinfo}[2]{#2}
\providecommand{\BIBentrySTDinterwordspacing}{\spaceskip=0pt\relax}
\providecommand{\BIBentryALTinterwordstretchfactor}{4}
\providecommand{\BIBentryALTinterwordspacing}{\spaceskip=\fontdimen2\font plus
\BIBentryALTinterwordstretchfactor\fontdimen3\font minus
  \fontdimen4\font\relax}
\providecommand{\BIBforeignlanguage}[2]{{%
\expandafter\ifx\csname l@#1\endcsname\relax
\typeout{** WARNING: IEEEtran.bst: No hyphenation pattern has been}%
\typeout{** loaded for the language `#1'. Using the pattern for}%
\typeout{** the default language instead.}%
\else
\language=\csname l@#1\endcsname
\fi
#2}}
\providecommand{\BIBdecl}{\relax}
\BIBdecl

\bibitem{shintake2018soft}
J.~Shintake, V.~Cacucciolo, D.~Floreano, and H.~Shea, ``Soft robotic
  grippers,'' \emph{Advanced Materials}, vol.~30, no.~29, p. 1707035, 2018.

\bibitem{xavier2022soft}
M.~S. Xavier, C.~D. Tawk, A.~Zolfagharian, J.~Pinskier, D.~Howard, T.~Young,
  J.~Lai, S.~M. Harrison, Y.~K. Yong, M.~Bodaghi \emph{et~al.}, ``Soft
  pneumatic actuators: A review of design, fabrication, modeling, sensing,
  control and applications,'' \emph{IEEE Access}, 2022.

\bibitem{manti2016stiffening}
M.~Manti, V.~Cacucciolo, and M.~Cianchetti, ``Stiffening in soft robotics: A
  review of the state of the art,'' \emph{IEEE Robotics \& Automation
  Magazine}, vol.~23, no.~3, pp. 93--106, 2016.

\bibitem{fitzgerald_review_2020}
\BIBentryALTinterwordspacing
S.~G. Fitzgerald, G.~W. Delaney, and D.~Howard, ``\BIBforeignlanguage{en}{A
  {Review} of {Jamming} {Actuation} in {Soft} {Robotics}},''
  \emph{\BIBforeignlanguage{en}{Actuators}}, vol.~9, no.~4, p. 104, Dec. 2020,
  number: 4 Publisher: Multidisciplinary Digital Publishing Institute.
  [Online]. Available: \url{https://www.mdpi.com/2076-0825/9/4/104}
\BIBentrySTDinterwordspacing

\bibitem{brown_universal_2010}
\BIBentryALTinterwordspacing
E.~Brown, N.~Rodenberg, J.~Amend, A.~Mozeika, E.~Steltz, M.~R. Zakin,
  H.~Lipson, and H.~M. Jaeger, ``\BIBforeignlanguage{en}{Universal robotic
  gripper based on the jamming of granular material},''
  \emph{\BIBforeignlanguage{en}{Proceedings of the National Academy of
  Sciences}}, vol. 107, no.~44, pp. 18\,809--18\,814, Nov. 2010, publisher:
  National Academy of Sciences Section: Physical Sciences. [Online]. Available:
  \url{https://www.pnas.org/content/107/44/18809}
\BIBentrySTDinterwordspacing

\bibitem{jiang2012design}
A.~Jiang, G.~Xynogalas, P.~Dasgupta, K.~Althoefer, and T.~Nanayakkara, ``Design
  of a variable stiffness flexible manipulator with composite granular jamming
  and membrane coupling,'' in \emph{2012 IEEE/RSJ International Conference on
  Intelligent Robots and Systems}.\hskip 1em plus 0.5em minus 0.4em\relax IEEE,
  2012, pp. 2922--2927.

\bibitem{fujita2016variable}
M.~Fujita, K.~Tadakuma, E.~Takane, T.~Ichimura, H.~Komatsu, A.~Nomura,
  M.~Konyo, and S.~Tadokoro, ``Variable inner volume mechanism for soft and
  robust gripping—improvement of gripping performance for large-object
  gripping,'' in \emph{2016 IEEE International Symposium on Safety, Security,
  and Rescue Robotics (SSRR)}.\hskip 1em plus 0.5em minus 0.4em\relax IEEE,
  2016, pp. 390--395.

\bibitem{amend_positive_2012}
J.~R. Amend, E.~Brown, N.~Rodenberg, H.~M. Jaeger, and H.~Lipson, ``A
  {Positive} {Pressure} {Universal} {Gripper} {Based} on the {Jamming} of
  {Granular} {Material},'' \emph{IEEE Transactions on Robotics}, vol.~28,
  no.~2, pp. 341--350, Apr. 2012.

\bibitem{delaney2012}
G.~W. Delaney, P.~W. Cleary, M.~Hilden, and R.~D. Morrison, ``Testing the
  validity of the spherical {{DEM}} model in simulating real granular screening
  processes,'' \emph{Chemical Engineering Science}, vol.~68, no.~1, pp.
  215--226, Jan. 2012.

\bibitem{nishida_development_2014}
T.~Nishida, D.~Shigehisa, N.~Kawashima, and K.~Tadakuma, ``Development of
  universal jamming gripper with a force feedback mechanism,'' in \emph{2014
  {Joint} 7th {International} {Conference} on {Soft} {Computing} and
  {Intelligent} {Systems} ({SCIS}) and 15th {International} {Symposium} on
  {Advanced} {Intelligent} {Systems} ({ISIS})}, Dec. 2014, pp. 242--246.

\bibitem{licht_universal_2016}
S.~Licht, E.~Collins, D.~Ballat-Durand, and M.~Lopes-Mendes, ``Universal
  jamming grippers for deep-sea manipulation,'' in \emph{{OCEANS} 2016
  {MTS}/{IEEE} {Monterey}}, Sep. 2016, pp. 1--5.

\bibitem{miettinen_granular_2019}
J.~Miettinen, P.~Frilund, I.~Vuorinen, P.~Kuosmanen, and P.~Kiviluoma,
  ``Granular jamming based robotic gripper for heavy objects,''
  \emph{Proceedings of the Estonian Academy of Sciences}, vol.~68, p. 421, Nov.
  2019.

\bibitem{howard2021shape}
D.~Howard, J.~O'Connor, J.~Brett, and G.~W. Delaney, ``Shape, size, and
  fabrication effects in 3d printed granular jamming grippers,'' in \emph{2021
  {IEEE} {International} {Conference} on {Soft} {Robotics} (Robosoft)}, 2021.

\bibitem{howard_one-shot_2021}
\BIBentryALTinterwordspacing
G.~D. Howard, J.~Brett, J.~O'Connor, J.~Letchford, and G.~W. Delaney,
  ``One-{Shot} {3D}-{Printed} {Multimaterial} {Soft} {Robotic} {Jamming}
  {Grippers},'' \emph{Soft Robotics}, Jun. 2021, publisher: Mary Ann Liebert,
  Inc., publishers. [Online]. Available:
  \url{https://www-liebertpub-com.ezproxy.library.uq.edu.au/doi/full/10.1089/soro.2020.0154}
\BIBentrySTDinterwordspacing

\bibitem{fitzgerald_evolving_2021}
\BIBentryALTinterwordspacing
S.~G. Fitzgerald, G.~W. Delaney, D.~Howard, and F.~Maire, ``Evolving soft
  robotic jamming grippers,'' in \emph{Proceedings of the {Genetic} and
  {Evolutionary} {Computation} {Conference}}, ser. {GECCO} '21.\hskip 1em plus
  0.5em minus 0.4em\relax New York, NY, USA: Association for Computing
  Machinery, Jun. 2021, pp. 102--110. [Online]. Available:
  \url{https://doi.org/10.1145/3449639.3459331}
\BIBentrySTDinterwordspacing

\bibitem{delaney_multi-objective_2020}
\BIBentryALTinterwordspacing
G.~W. Delaney and G.~Howard, ``Multi-objective exploration of a granular matter
  design space,'' in \emph{Proceedings of the 2020 {Genetic} and {Evolutionary}
  {Computation} {Conference} {Companion}}, ser. {GECCO} '20.\hskip 1em plus
  0.5em minus 0.4em\relax New York, NY, USA: Association for Computing
  Machinery, Jul. 2020, pp. 263--264. [Online]. Available:
  \url{https://doi.org/10.1145/3377929.3389951}
\BIBentrySTDinterwordspacing

\bibitem{chopra_granular_2020}
S.~Chopra, M.~T. Tolley, and N.~Gravish, ``Granular {Jamming} {Feet} {Enable}
  {Improved} {Foot}-{Ground} {Interactions} for {Robot} {Mobility} on
  {Deformable} {Ground},'' \emph{IEEE Robotics and Automation Letters}, vol.~5,
  no.~3, pp. 3975--3981, Jul. 2020, conference Name: IEEE Robotics and
  Automation Letters.

\bibitem{cheng_prosthetic_2016}
\BIBentryALTinterwordspacing
N.~Cheng, J.~Amend, T.~Farrell, D.~Latour, C.~Martinez, J.~Johansson,
  A.~McNicoll, M.~Wartenberg, S.~Naseef, W.~Hanson, and W.~Culley, ``Prosthetic
  {Jamming} {Terminal} {Device}: {A} {Case} {Study} of {Untethered} {Soft}
  {Robotics},'' \emph{Soft Robotics}, vol.~3, no.~4, pp. 205--212, Dec. 2016,
  publisher: Mary Ann Liebert, Inc., publishers. [Online]. Available:
  \url{https://www.liebertpub.com/doi/abs/10.1089/soro.2016.0017}
\BIBentrySTDinterwordspacing

\bibitem{zubrycki_novel_2017}
\BIBentryALTinterwordspacing
I.~Zubrycki and G.~Granosik, ``\BIBforeignlanguage{en}{Novel {Haptic} {Device}
  {Using} {Jamming} {Principle} for {Providing} {Kinaesthetic} {Feedback} in
  {Glove}-{Based} {Control} {Interface}},''
  \emph{\BIBforeignlanguage{en}{Journal of Intelligent \& Robotic Systems}},
  vol.~85, no.~3, pp. 413--429, Mar. 2017. [Online]. Available:
  \url{https://doi.org/10.1007/s10846-016-0392-6}
\BIBentrySTDinterwordspacing

\bibitem{kapadia_design_2012}
J.~Kapadia and M.~Yim, ``Design and performance of nubbed fluidizing jamming
  grippers,'' in \emph{2012 {IEEE} {International} {Conference} on {Robotics}
  and {Automation}}, May 2012, pp. 5301--5306, iSSN: 1050-4729.

\bibitem{amend_soft_2016}
J.~Amend, N.~Cheng, S.~Fakhouri, and B.~Culley, ``\BIBforeignlanguage{eng}{Soft
  {Robotics} {Commercialization}: {Jamming} {Grippers} from {Research} to
  {Product}},'' \emph{\BIBforeignlanguage{eng}{Soft Robotics}}, vol.~3, no.~4,
  pp. 213--222, Dec. 2016.

\bibitem{janda_unjamming_2009}
\BIBentryALTinterwordspacing
A.~Janda, D.~Maza, A.~Garcimartín, E.~Kolb, J.~Lanuza, and E.~Clément,
  ``\BIBforeignlanguage{en}{Unjamming a granular hopper by vibration},''
  \emph{\BIBforeignlanguage{en}{EPL (Europhysics Letters)}}, vol.~87, no.~2, p.
  24002, Aug. 2009, publisher: IOP Publishing. [Online]. Available:
  \url{https://iopscience.iop.org/article/10.1209/0295-5075/87/24002/meta}
\BIBentrySTDinterwordspacing

\bibitem{nicodemi_compaction_1997}
\BIBentryALTinterwordspacing
M.~Nicodemi, A.~Coniglio, and H.~J. Herrmann,
  ``\BIBforeignlanguage{en}{Compaction and force propagation in granular
  packings},'' \emph{\BIBforeignlanguage{en}{Physica A: Statistical Mechanics
  and its Applications}}, vol. 240, no.~3, pp. 405--418, Jun. 1997. [Online].
  Available:
  \url{http://www.sciencedirect.com/science/article/pii/S0378437197001623}
\BIBentrySTDinterwordspacing

\bibitem{nicodemi_frustration_1997}
M.~Nicodemi, A.~Coniglio, and H.~Herrmann, ``Frustration and slow dynamics of
  granular packing,'' \emph{Physical Review E - PHYS REV E}, vol.~55, pp.
  3962--3969, Apr. 1997.

\bibitem{windows-yule_resonance_2015}
\BIBentryALTinterwordspacing
C.~R.~K. Windows-Yule, A.~D. Rosato, A.~R. Thornton, and D.~J. Parker,
  ``Resonance effects on the dynamics of dense granular beds: achieving optimal
  energy transfer in vibrated granular systems,'' \emph{New Journal of
  Physics}, vol.~17, no.~2, p. 023015, Feb. 2015. [Online]. Available:
  \url{https://iopscience.iop.org/article/10.1088/1367-2630/17/2/023015}
\BIBentrySTDinterwordspacing

\bibitem{lemrich_dynamic_2017}
\BIBentryALTinterwordspacing
L.~Lemrich, J.~Carmeliet, P.~A. Johnson, R.~Guyer, and X.~Jia, ``Dynamic
  induced softening in frictional granular materials investigated by
  discrete-element-method simulation,'' \emph{Physical Review E}, vol.~96,
  no.~6, p. 062901, Dec. 2017, publisher: American Physical Society. [Online].
  Available: \url{https://link.aps.org/doi/10.1103/PhysRevE.96.062901}
\BIBentrySTDinterwordspacing

\bibitem{chen_low-frequency_2010}
\BIBentryALTinterwordspacing
K.~Chen, W.~G. Ellenbroek, Z.~Zhang, D.~T.~N. Chen, P.~J. Yunker, S.~Henkes,
  C.~Brito, O.~Dauchot, W.~van Saarloos, A.~J. Liu, and A.~G. Yodh,
  ``Low-{Frequency} {Vibrations} of {Soft} {Colloidal} {Glasses},''
  \emph{Physical Review Letters}, vol. 105, no.~2, p. 025501, Jul. 2010,
  publisher: American Physical Society. [Online]. Available:
  \url{https://link.aps.org/doi/10.1103/PhysRevLett.105.025501}
\BIBentrySTDinterwordspacing

\bibitem{delaney2010g}
G.~W. Delaney and P.~W. Cleary, ``The packing properties of superellipsoids,''
  \emph{EPL (Europhysics Letters)}, vol.~89, no.~3, p. 34002, 2010.

\end{thebibliography}

\appendix
\section{Statistical Testing}
Statistical tests for all experiments. Mann-Whitney U tests were conducted for all pairs of conditions within an experiment.  Top 3: frequency tests, bottom 3: volume tests.

\begin{figure}[h!]
\centering
\includegraphics[width=0.6\columnwidth]{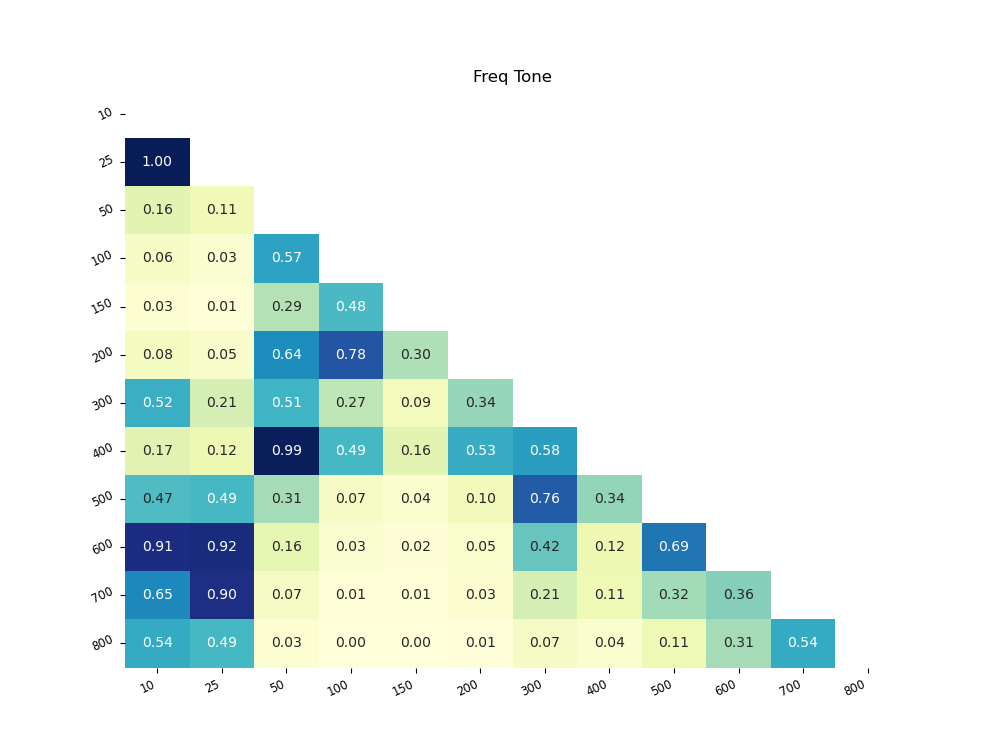}\\
\includegraphics[width=0.6\columnwidth]{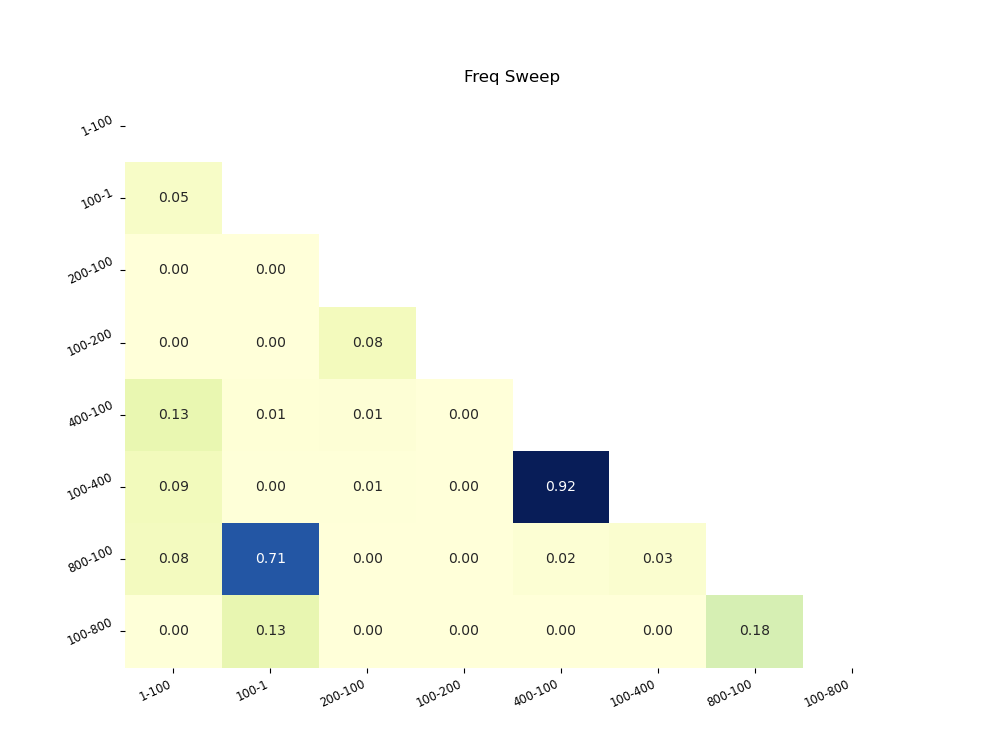}\\
\includegraphics[width=0.6\columnwidth]{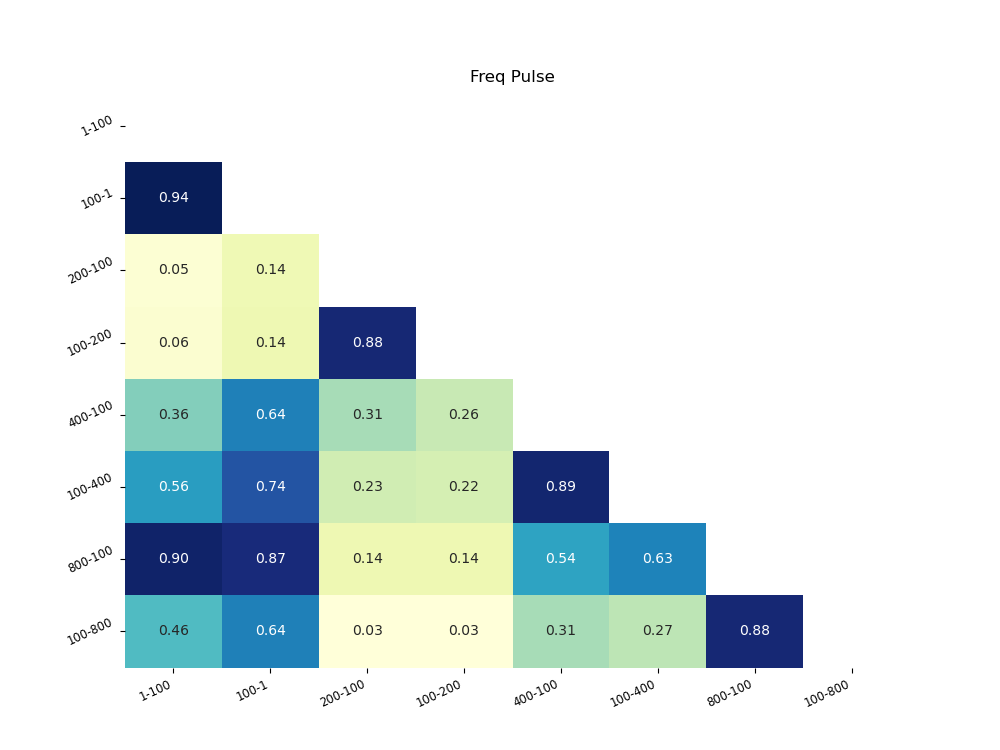}
\label{fig:appendix_freq_stat_tests}
\caption{Frequency Mann-Whitney summary.}
\end{figure} 

\begin{figure}[h!]
\centering
\includegraphics[width=0.6\columnwidth]{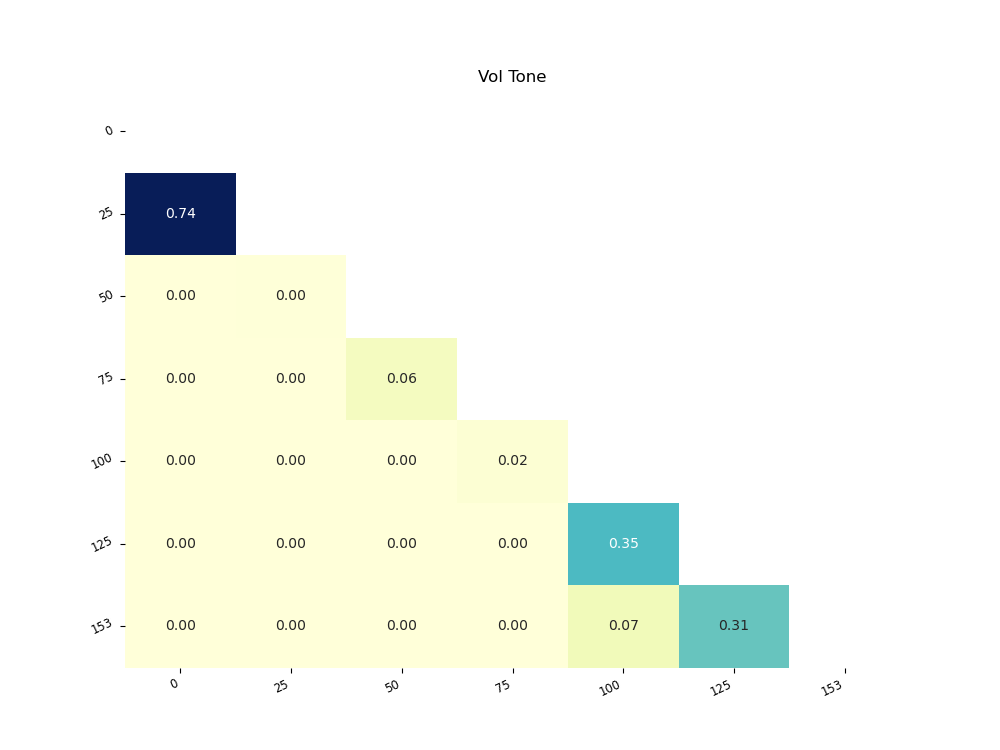}\\
\includegraphics[width=0.6\columnwidth]{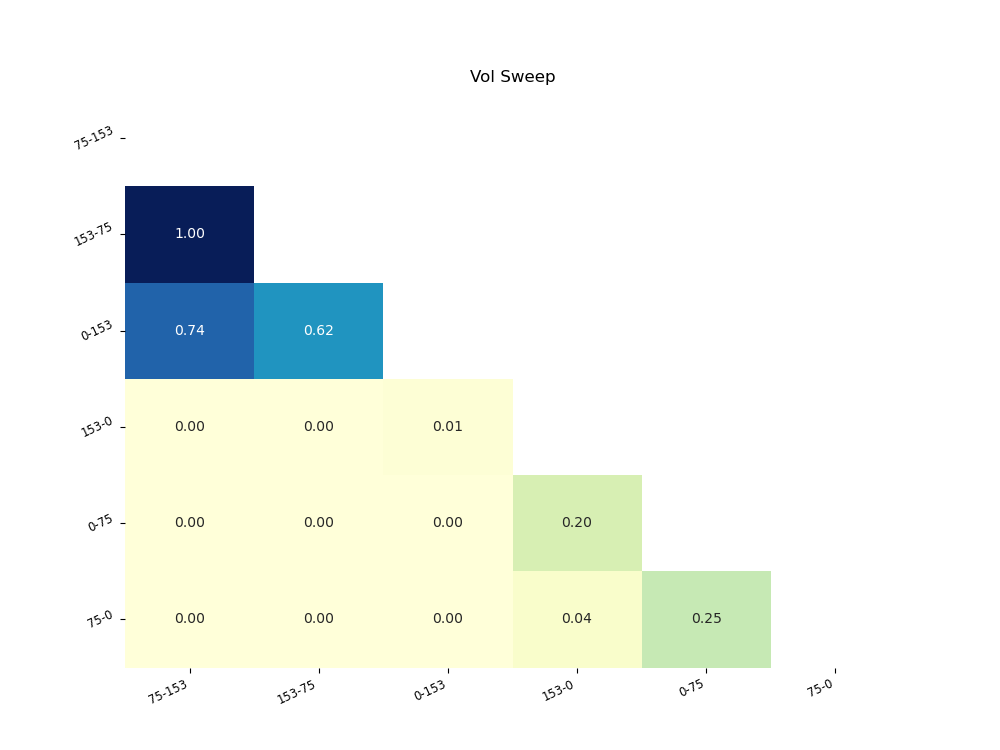}\\
\includegraphics[width=0.6\columnwidth]{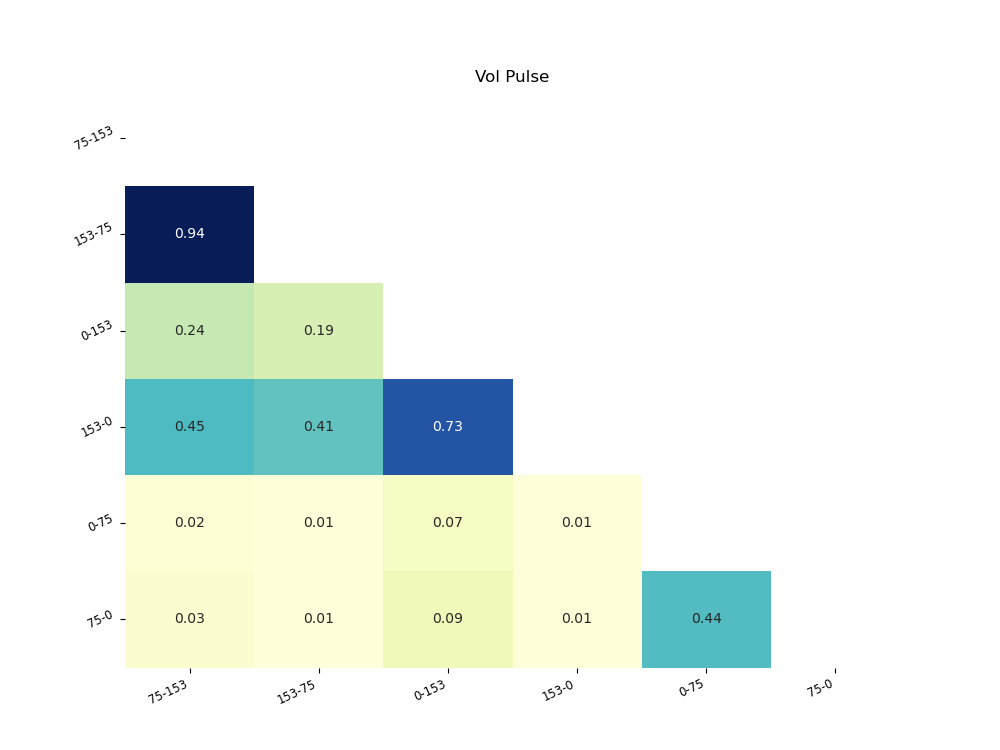}
\label{fig:appendix_vol_stat_tests}
\caption{Volume Mann-Whitney summary.}
\end{figure} 

\end{document}